\documentclass[letterpaper,10pt,conference]{vendor/ieeeconf}

\IEEEoverridecommandlockouts
\ifdefined\pdfminorversion
\fi

\usepackage[T1]{fontenc}
\usepackage[utf8]{inputenc}
\usepackage{amsmath,amssymb}
\usepackage{booktabs}
\usepackage{array}
\usepackage{multirow}
\usepackage{graphicx}
\usepackage{xcolor}
\usepackage{microtype}
\usepackage{url}
\usepackage{balance}
\usepackage{tikz}
\usetikzlibrary{arrows.meta,positioning,fit,calc,shapes.geometric}

\definecolor{jepablue}{HTML}{1769AA}
\definecolor{jepalight}{HTML}{E8F2FA}
\definecolor{targetorange}{HTML}{E8872D}
\definecolor{targetlight}{HTML}{FCEEDF}
\definecolor{goodgreen}{HTML}{2A8F5B}
\definecolor{badred}{HTML}{B8433F}
\definecolor{neutralgray}{HTML}{6A7178}

\ifdefined
\fi

\newcommand{\candidate}{CS-JEPA}
\newcommand{\recon}{Future-Recon}
\newcommand{\collectiveprediction}{\texttt{collective\_prediction}}

\title{\LARGE \bf
One Future, Every Robot: Label-Efficient Collective-State Prediction with Decentralized JEPA
}

\author{Alan-Barsag Gazzaev, Alexey Gavrilov, and Sergey Muravyov%
\thanks{The authors are with ITMO University, Saint Petersburg, Russia.}}

\begin{document}

\maketitle
\thispagestyle{empty}
\pagestyle{empty}

\begin{abstract}
Decentralized robots often need a common view of what their team is becoming, even though each
robot sees different evidence and cannot rely on a central estimate or output-level consensus.  We
ask whether compatible collective-state predictions can emerge under this constraint.
Collective-State JEPA (CS-JEPA) trains every robot to predict the same fixed-width latent future
from its own history and bounded neighbor messages, with no agreement loss; predictions and plans
are never pooled at deployment.  In a fresh independent replication, agreement improves for every
seed and every evaluated split.  Accuracy improves at the same time, ruling out the uninformative
solution in which all robots merely collapse to one prediction: relative to capacity-matched
raw-future reconstruction, collective-state error falls by 28.4\% in distribution and by
64.4--75.6\% under topology and swarm-size shift.  Translation-free and crossed-pretraining
controls preserve this joint result, while action-conditioned and rigid-body evaluations show that
the receiver-local representation supports independent decisions.  A shared latent future can
therefore align decentralized predictions without consensus training while preserving useful,
label-efficient information.
\end{abstract}

%%%%%%%%%%%%%%%%%%%%%%%%%%%%%%%%%%%%%%%%%%%%%%%%%%%%%%%%%%%%%%%%%%%%%%%%%%%%%%%%
\section{Introduction}

A robot swarm can flock, fragment, rotate, or lose connectivity even though no member sees that
global regime directly.  Each robot receives only a fragment of the evidence, so independently
plausible predictions can still describe incompatible futures.  This is a distinctly decentralized
failure mode: useful coordination requires a common view of what the formation is becoming, but
there is no central estimate to copy back and no opportunity to pool robot outputs.  The problem is
therefore neither ordinary per-agent trajectory forecasting nor a centralized critic state.  Every
receiver must infer a compatible future collective quantity from a different local history.

Two research threads approach complementary halves of this problem.  Permutation-invariant
encoders, graph policies, and global-state predictors show how local exchange can expose hidden
swarm structure~\cite{zaheer2017deepsets,huttenrauch2019swarm,tolstaya2020gnn,bloom2023gsp,bloom2025gspn}.
Multi-agent representation learning and world models use prediction to improve communication and
control~\cite{guan2022masia,feng2025timar,zhang2025dtca,nomura2025dcwm}, while JEPAs replace raw
reconstruction with prediction in a learned target space~\cite{lecun2022path,assran2023ijepa}.
What remains unresolved is their conjunction: can a latent target represent the future
\emph{swarm as a set}, bring independent receiver-local predictions into agreement without a
consensus loss, and remain useful when topology, swarm size, and downstream label budget change?

CS-JEPA starts from one hypothesis: training different local views toward the same latent future
should align what those views mean, even when robots are never trained to match one another's
outputs.  Compatibility at zero agreement-loss weight is the most direct test of that hypothesis.
Agreement alone, however, would be weak evidence---constant predictions agree perfectly.  The
critical result must therefore be joint: independently formed estimates should become more
compatible while retaining, and ideally improving, their physical accuracy.  Label efficiency,
topology and size transfer, and downstream use then test whether this alignment preserves useful
collective dynamics rather than merely suppressing receiver-specific variation.

\begin{figure*}[t]
\centering
\begin{tikzpicture}[
    font=\footnotesize,
    >=Latex,
    box/.style={draw,rounded corners=2pt,minimum height=8mm,align=center,inner sep=3pt},
    bluebox/.style={box,draw=jepablue,fill=jepalight},
    orangebox/.style={box,draw=targetorange,fill=targetlight},
    graybox/.style={box,draw=neutralgray!75,fill=neutralgray!8},
    arr/.style={->,line width=.65pt,draw=neutralgray!90},
    msg/.style={->,line width=.8pt,draw=jepablue},
    train/.style={->,line width=.7pt,dashed,draw=targetorange}
]
% Deployment path
\node[font=\bfseries,anchor=west,text=jepablue] at (-8.2,1.25)
{DEPLOYMENT: the same receiver-local computation at every robot};
\node[bluebox,minimum width=2.35cm] (hist) at (-6.75,.35) {16-frame local\\history at robot $i$};
\node[bluebox,minimum width=2.35cm] (memory) at (-3.8,.35) {synchronous GRU\\$h_t^i\in\mathbb{R}^{64}$};
\node[bluebox,minimum width=2.15cm] (pred) at (-.95,.35) {shared-target\\predictor};
\node[orangebox,minimum width=2.9cm] (out) at (2.05,.35)
{every $i$: $\widehat Z_{t+4}^{,i}$\\same 17 $\times$ 65 target};
\node[orangebox,minimum width=2.1cm] (probe) at (5.55,.35) {small labeled\\probe};
\draw[arr] (hist) -- (memory);
\draw[arr] (memory) -- node[above=2pt]{$h_t^i$} (pred);
\draw[arr] (pred) -- (out);
\draw[arr] (out) -- node[below=2pt,align=center,font=\scriptsize]{6/12/24\\labels} (probe);
\node[bluebox,minimum width=2.5cm] (neighbors) at (-3.8,-.75)
{mean neighbor memory\\$\bar h_{t-1}^{,i}$};
\draw[msg] (neighbors.north) -- (memory.south);
\node[align=center,text=neutralgray] at (-6.75,-.75)
{no global pool; no clock\\zero action inputs};
\node[align=center,text=jepablue] at (-1.05,-.75)
{one message round\\256 B/edge/step};

\node[orangebox,minimum width=2.45cm,minimum height=6mm,inner sep=2pt]
(loss) at (2.05,-1.15) {JEPA target loss\\to every receiver};

% Training-only path
\node[font=\bfseries,anchor=west,text=targetorange] at (-8.2,-1.85) {PRETRAINING ONLY};
\node[graybox,minimum width=2.55cm] (future) at (-6.75,-2.75)
{privileged swarm state\\at $t{+}2,t{+}4$};
\node[graybox,minimum width=2.55cm] (targetenc) at (-3.8,-2.75)
{frozen target encoder\\+ set tokenizer};
\node[orangebox,minimum width=2.6cm] (target) at (-.75,-2.75)
{one $Z_{t+h}$\\global $+$ $4\!\times\!4$ field};
\node[graybox,minimum width=2.85cm] (anchor) at (2.35,-2.75)
{$+\ \lambda=2$ receiver anchor\\training-only head};
\node[draw=badred,rounded corners=2pt,inner sep=3pt,align=center,text=badred,minimum width=2.5cm]
(removed) at (5.65,-2.75) {removed at runtime:\\target path and anchor};
\draw[train] (future) -- (targetenc);
\draw[train] (targetenc) -- (target);
\draw[train] (target) -- (anchor);
\draw[train] (anchor) -- (removed);
\draw[train] (target.north) -- (loss.south west);
\draw[train] (loss.north) -- (out.south);
\end{tikzpicture}
\caption{\textbf{One future, every robot.}  At deployment, robot $i$ uses only its local history and
received recurrent memories to output $\widehat Z_{t+4}^{,i}$; robot $j$ independently predicts the
same shared target.  The privileged future path and receiver anchor exist only during pretraining.
The reconstruction reference keeps the blue path, output width, and identical receiver anchor, but
replaces latent-target prediction with a training-only raw-state decoder.}
\label{fig:method}
\end{figure*}
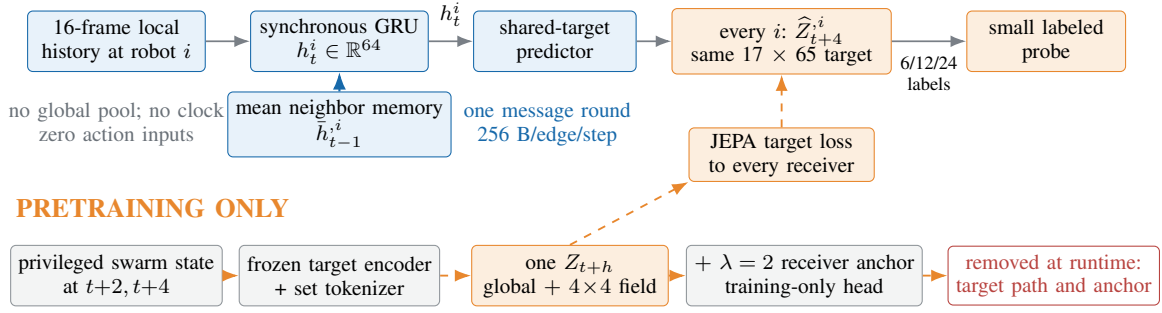

Figure~\ref{fig:method} summarizes the resulting architecture.  Each robot recurrently transports
local evidence and predicts a fixed global-plus-spatial token field at two future horizons.  A small
probe decodes collective variables such as polarization, connectivity, and task score.  Obtaining
these labels requires whole-swarm sensing or centralized post-processing during evaluation, while
deployed robots use neither.  Low-label probing therefore asks whether physically meaningful state
is readily accessible from the learned future.  Our experiments follow one scientific chain:
compatible estimates should emerge without consensus training, remain accurate with scarce labels
under distribution shift, and retain information that matters for decentralized decisions.

Our contributions are:
\begin{itemize}
    \item a decentralized shared-state prediction problem and a size-invariant common-future JEPA
    whose recurrent communication and output width remain fixed as the swarm grows;
    \item fresh-cohort evidence that receiver-local predictions become more compatible without an
    agreement loss while accuracy improves across label budgets, topology shifts, and swarm-size
    shifts, supported by translation-free and crossed-objective controls; and
    \item action-conditioned value estimation and fully decentralized rigid-body control showing
    that the learned common future remains useful when each robot selects and executes its own plan.
\end{itemize}

%%%%%%%%%%%%%%%%%%%%%%%%%%%%%%%%%%%%%%%%%%%%%%%%%%%%%%%%%%%%%%%%%%%%%%%%%%%%%%%%
\section{Related Work}

\subsection{Global information from local interactions}

A recurring idea in swarm learning is that global structure can be assembled from local exchange.
Mean embeddings, invariant set functions, and graph neural networks exploit agent
exchangeability~\cite{zaheer2017deepsets,huttenrauch2019swarm,tolstaya2020gnn}.  Otte's artificial
group mind distributes one wireless network across a physical swarm to classify present
environmental patterns~\cite{otte2018groupmind}, while distributed spatial awareness builds a
shared present-day frame through local factor graphs~\cite{jones2025dsa}.  Global State Prediction
(GSP) moves the question into the future for policy learning~\cite{bloom2023gsp}, and GSP-N adds
bandwidth-limited communication for collective transport~\cite{bloom2025gspn}.  Together these
works show that local communication can recover useful global context.  Our question is narrower
and directly measurable: can every receiver independently reach a compatible estimate of one
future collective target, while remaining accurate with few globally labeled episodes?

\subsection{Predictive representations for multiple agents}

Multi-agent representation learning asks how prediction can make a team act better.  MASIA
self-supervises a compact message aggregate by reconstructing and predicting future
information~\cite{guan2022masia}; TIMAR uses a joint transition model to inform local
representations~\cite{feng2025timar}.  Recent world models combine decentralized dynamics with
centralized aggregation~\cite{zhang2025dtca}, separate local and joint latent states
~\cite{xue2026dmawm}, or learn communication and coordination directly~\cite{nomura2025dcwm}.
These systems are usually judged by return, imagined dynamics, or emergent communication.  We
instead expose the shared future as an endpoint in its own right: $N$ receiver-local estimates of
one swarm-level quantity, measured for both accuracy and agreement before they are used for control.

\subsection{JEPA beyond single observations}

JEPA is a natural candidate for this endpoint because it predicts what matters in a learned target
space instead of reconstructing every observation detail~\cite{lecun2022path,assran2023ijepa,bardes2024vjepa}.
Population-level JEPA shows that a set can itself be the learning object, although in a centralized
biological setting~\cite{bakulin2026population}.  V2X-JEPA brings joint-embedding prediction to
cooperative perception and studies annotation efficiency and communication disruption
~\cite{mayumu2026v2xjepa}; TrajJEPA targets multi-agent trajectory forecasting
~\cite{yang2026trajjepa}.  These works support latent prediction across multiple entities, but their
outputs are fused detections, trajectories, or centralized population representations.  CS-JEPA
tests the complementary case in which every robot must independently predict the same
size-invariant future swarm state through a fixed decentralized interface.

%%%%%%%%%%%%%%%%%%%%%%%%%%%%%%%%%%%%%%%%%%%%%%%%%%%%%%%%%%%%%%%%%%%%%%%%%%%%%%%%
\section{Problem Formulation}

The challenge is not to define a global future; a simulator or centralized training system can do
that directly.  The challenge is to make that future predictable from many different local
information sets without copying a central estimate back to the robots.  At time $t$, the active
swarm is a dynamic graph
$\mathcal G_t=(\mathcal V_t,\mathcal E_t)$.  Robot $i$ observes a receiver-indexed local view
$o_t^i$ and receives messages only from $\mathcal N_t(i)$.  Its allowed information is
\begin{equation}
\mathcal I_t^i=\left\{o_{t-15:t}^i,\;\mathcal N_{t-15:t}(i),\;
\left(m_{\tau}^{j\rightarrow i}\right)_{
\substack{\tau=t-15:t\\j\in\mathcal N_\tau(i)}}\right\}.
\label{eq:info}
\end{equation}
Here $o_t^i$ contains only robot $i$'s normalized world position, velocity, two-dimensional task
vector, and active bit, all in a shared normalized world frame.  The task vector is the desired
heading $(1,0)$ for flocking and the world-size-normalized displacement from robot $i$ to its
assigned target for formation and coverage.  Raw neighbor features are unavailable: neighbors
contribute only their previous 64-float recurrent memories.  The normalized episode clock is
zeroed and recorded future actions are not provided.  We seek a
single permutation-invariant future target $Z_{t+h}=T_\xi(S_{t+h})$ for the whole active set, but a
distinct prediction is produced at every robot:
\begin{equation}
\widehat Z_{t+h}^{,i}=P_\theta(\mathcal I_t^i), \qquad
i\in\mathcal V_t,\quad h\in\{2,4\}.
\label{eq:goal}
\end{equation}
The target is common, but the evidence and prediction error remain receiver-specific.  We call the
final $h=4$ output \collectiveprediction.  No average over robot predictions is used as the
primary representation: success means that each robot can construct the shared future for itself.

For downstream evaluation, a probe $g_\phi$ maps each frozen $\widehat Z_{t+4}^{,i}$ to ten
future collective quantities: polarization, target alignment, cohesion, dispersion, angular
momentum, connectivity, normalized cluster count, collision fraction, mean speed, and task score.
Thus, ``collective state'' denotes a learned global-plus-spatial latent target with an explicitly
audited physical decoding task; it is not the concatenated state vector of all robots.

\subsection{Decentralization and scaling semantics}

Equation~\eqref{eq:goal} defines $|\mathcal V_t|$ outputs, not one estimate copied back from a
central node.  Robots share parameters and target semantics, but robot $i$ retains its own hidden
state and never reads robot $j$'s prediction.  Information can travel beyond one hop only through
successive recurrent updates, so a 16-frame context is not equivalent to instantaneous global
access.  Agreement is therefore a substantive result: when accuracy improves at the same time,
compatible predictions cannot be attributed to a shared output or explicit consensus step.

For a directed communication graph, the prescribed traffic at one step is
$256|\mathcal E_t|$ bytes in total and $256|\mathcal N_t(i)|$ bytes received by robot $i$.
Mean aggregation fixes the tensor width presented to the GRU as degree changes; the tokenizer
similarly fixes the output width as $N$ changes.  These choices make model parameters independent
of swarm size and keep per-robot traffic bounded on the degree-four training graphs.  They do not
make total network traffic independent of $N$, nor do they provide a centralized shortcut.

%%%%%%%%%%%%%%%%%%%%%%%%%%%%%%%%%%%%%%%%%%%%%%%%%%%%%%%%%%%%%%%%%%%%%%%%%%%%%%%%
\section{Collective-State JEPA}

\subsection{Receiver-local recurrent transport}

The communication module should carry evidence, not a centralized answer.  A frozen local encoder
maps robot $i$'s self observation to $e_t^i\in\mathbb R^{64}$, and each active robot broadcasts
only its previous recurrent state.  Receiver $i$ computes
\begin{align}
\bar h_{t-1}^{,i} &= \frac{1}{|\mathcal N_t(i)|}
\sum_{j\in\mathcal N_t(i)} h_{t-1}^j, \\
h_t^i &= \operatorname{GRU}([e_t^i,\bar h_{t-1}^{,i}],h_{t-1}^i).
\label{eq:gru}
\end{align}
We set $\bar h_{t-1}^{,i}=0$ when $\mathcal N_t(i)=\varnothing$, matching the masked-mean
implementation.  Mean aggregation keeps the input width fixed as degree changes, while recurrence
lets evidence travel over several steps without exposing another robot's prediction.  Updates are
synchronous, use one message round per environment step, and send 64 float32 values (256 bytes)
per directed edge.  With 16 frames, the measured warm-up is 15 message steps.  There is no
memory-consensus mixing and no global readout at inference.

\subsection{One variable-size future target}

A single global vector would ignore spatial layout, whereas one token per robot would grow with the
swarm.  We use a middle ground: the frozen target encoder embeds every active future robot, and a
permutation-invariant tokenizer forms one global token plus a fixed $4\times4$ spatial field.  At
anchor $a_k$,
\begin{equation}
\begin{aligned}
z_k&=\left[
\frac{\sum_j w_{jk}e_{t+h}^j}{\sum_j w_{jk}},\;
\frac{\sum_j w_{jk}}{|\mathcal V_{t+h}|}
\right],\\[-1mm]
w_{jk}&=\exp\!\left[-\frac{\|p_j-a_k\|^2}{2\sigma^2}\right].
\end{aligned}
\label{eq:token}
\end{equation}
with $\sigma=0.22$.  The global token uses the active-set mean embedding and a presence mass.
Seventeen 65-D tokens produce a fixed 1105-D $Z_{t+h}$ regardless of swarm size.  Token roles are
fixed; future graph adjacency is not encoded into the target.

A role-conditioned predictor recursively forecasts $t{+}2$ then $t{+}4$, stopping the gradient
through its first prediction.  Let $\mathcal B$ be a minibatch,
$\mathcal H=\{2,4\}$, $D_Z=1105$, and
$M_Z=\max\{1,|\mathcal H|\sum_{b\in\mathcal B}|\mathcal V_t^b|\}$.
The implemented common-target reduction is
\begin{equation}
\mathcal L_{\mathrm{CS}}=\frac{1}{M_Z}
\sum_{\substack{b\in\mathcal B,\;i\in\mathcal V_t^b\\h\in\mathcal H}}
\frac{\|\widehat Z_{t+h}^{b,i}-\operatorname{sg}(Z_{t+h}^{b})\|_2^2}{D_Z}.
\label{eq:jepa}
\end{equation}
Every receiver is trained toward the same target, but receivers are never trained toward one
another.  Agreement is therefore intentionally left out of the objective: its loss weight is zero.

\subsection{Receiver anchor and matched reconstruction}

Predicting only a set target can discard the receiver's own dynamics.  A training-only MLP
therefore predicts that receiver's frozen future embedding $e_{t+h}^{b,i}$ from $h_t^{b,i}$.
With $\mathcal Q=\{(b,i,h):b\in\mathcal B,\;h\in\mathcal H,\;
i\in\mathcal V_t^b\cap\mathcal V_{t+h}^b\}$, $D_e=64$, and
$M_e=\max\{1,|\mathcal Q|\}$, the implemented objective is
\begin{align}
\mathcal L_{\mathrm{anchor}}
&=\frac{1}{M_e}\sum_{(b,i,h)\in\mathcal Q}
\frac{\|\widehat e_{t+h}^{b,i}-\operatorname{sg}(e_{t+h}^{b,i})\|_2^2}{D_e}, \\
\mathcal L_{\mathrm{CS\text{-}JEPA}}
&=\mathcal L_{\mathrm{CS}}+\lambda\mathcal L_{\mathrm{anchor}},
\qquad \lambda=2.
\label{eq:anchor}
\end{align}
We use the same $\lambda=2$ anchor weight for both methods.  The anchor head is absent from
\texttt{predict\_collective} and deployment counts.

The primary comparison asks whether any gain comes from the common latent target rather than from
the recurrent architecture.  \recon{} therefore preserves the frozen encoder, messages,
role-conditioned predictor, 1105-D bottleneck, optimization schedule, probe, deployment parameter
count, and the same $\lambda=2$ receiver anchor.  Only the self-supervised target changes: a
training-only decoder reconstructs a global-plus-spatial field of normalized future position
$(x,y)$, velocity $(v_x,v_y)$, and task vector $(q_x,q_y)$, with the same relative-density mass
coordinate.  Paired methods begin from bitwise-identical shared-state initialization.  CS-JEPA has
139,137 online trainable pretraining parameters; \recon{} has 148,744 because its training-only raw
decoder adds 9,607.  Both deploy 123,713 representation parameters (134,773 with the probe), so
the comparison does not favor CS-JEPA through extra capacity.
Writing the decoded raw field as $\widehat R_{t+h}^{b,i}$ and its target as
$R_{t+h}^{b}\in\mathbb R^{D_R}$, $D_R=17(6+1)=119$, its matched objective is
\begin{equation}
\mathcal L_{\mathrm{FR}}=\frac{1}{M_Z}
\sum_{\substack{b\in\mathcal B,\;i\in\mathcal V_t^b\\h\in\mathcal H}}
\frac{\|\widehat R_{t+h}^{b,i}-R_{t+h}^{b}\|_2^2}{D_R}
+\lambda\mathcal L_{\mathrm{anchor}}.
\label{eq:recon}
\end{equation}

\subsection{Training-to-deployment separation}

Privileged future state acts as a teacher, not as a runtime input.  During pretraining, the online
branch is unrolled independently for every receiver using only \eqref{eq:info}; the target branch
computes $Z_{t+h}$ once from the future active set and supplies the same stopped-gradient target to
every receiver.  Future state never enters the GRU, messages, or predictor.  The receiver-anchor
head likewise contributes only a training loss and is never concatenated to
\collectiveprediction.  The privileged path therefore defines what should be predicted without
solving the prediction at inference time.

After checkpoint selection, we discard the target encoder, tokenizer, reconstruction decoder, and
anchor head.  The local encoder and recurrent predictor are frozen, and the ridge probe is fit from
receiver-local $h=4$ predictions.  Every robot runs the same function but carries its own recurrent
state and incurs its own error.  What is shared is the meaning of the target, not a synchronized
latent or server-produced estimate.  Table~\ref{tab:contract} summarizes this deployment boundary.

\begin{table}[!b]
\caption{Fixed representation and deployment contract.}
\label{tab:contract}
\centering
\footnotesize
\setlength{\tabcolsep}{3.2pt}
\begin{tabular}{p{.39\columnwidth}p{.52\columnwidth}}
\toprule
Component & Setting \\
\midrule
Local temporal context & 16 frames, clock zeroed, actions zero \\
Prediction horizons & $t{+}2$ and $t{+}4$ \\
Shared target & 1 global $+$ $4\!\times\!4$ spatial tokens \\
Target width & 65 per token; 1105 total \\
Recurrent message & 64 float32 values from previous step \\
Communication & 1 synchronous round; 256 B/edge/step \\
Message warm-up & 15 steps before a full 16-frame context \\
Deployment size & 123,713 representation parameters \\
With ten-output probe & 134,773 parameters \\
Privileged target path & pretraining only; absent at inference \\
\bottomrule
\end{tabular}
\end{table}

%%%%%%%%%%%%%%%%%%%%%%%%%%%%%%%%%%%%%%%%%%%%%%%%%%%%%%%%%%%%%%%%%%%%%%%%%%%%%%%%
\section{Experimental Setup}

\subsection{Data and training}

We separate two reasons a decentralized representation might fail to transfer: the communication
graph may change, or the swarm may simply become larger.  The label-efficiency evaluation covers
flocking, formation, and coverage in 200-step episodes with process noise 0.02 and no observation
noise.  Training and ID use bounded-small-world graphs with $N\in\{10,18\}$ and maximum degree
four.  Ring and mutual-$k$NN isolate topology shift; the size split uses unseen
$N\in\{36,72,108\}$, up to $6\times$ the largest training swarm (Table~\ref{tab:data}).

\begin{table}[!b]
\caption{Label-efficiency data family.}
\label{tab:data}
\centering
\footnotesize
\setlength{\tabcolsep}{3.2pt}
\begin{tabular}{lccc}
\toprule
Split & Graph & Robots & Episodes \\
\midrule
Train / label pool & small world & 10, 18 & 120 \\
Model selection only & small world & 10, 18 & 30 \\
ID test & small world & 10, 18 & 30 \\
Topology OOD & ring & 10, 18 & 30 \\
Topology OOD & mutual-$k$NN & 10, 18 & 30 \\
Size OOD & small world & 36, 72, 108 & 27 \\
\bottomrule
\end{tabular}
\end{table}

Both methods train for 50 epochs with batch size 256, Adam learning rate $5\!\times\!10^{-4}$,
cosine weight decay $10^{-7}\!\rightarrow\!10^{-6}$, and gradient clipping at 1.0.  Hidden and
latent widths are 128 and 64.  Within each anchor-matched cohort, both methods share the same
frozen stage-0 encoder; the initial and fresh cohorts contain five and twelve independently trained
stage-1 outer seeds, respectively.  Stage~0 is our separately trained mean-target
CS-JEPA encoder on unlabeled swarm episodes, not an external pretrained model; both its context and
target copies are loaded from the same checkpoint and remain frozen.  The crossed control below
separately retrains stage~0 under both latent-mean and raw-future objectives.  For each method and
seed, checkpoint selection uses only the model-selection split.

\subsection{Matched comparison and leakage controls}

The central comparison is designed to isolate the prediction target.  CS-JEPA and \recon{} use the
same unlabeled pretraining episodes, 16-frame receiver histories, recurrent backbone, message
payload, optimizer schedule,
checkpoint-selection split, label subsets, ridge probe, and test episodes.  The representation
width, receiver anchor, and deployed parameter count are also equal.  They differ in the primary
self-supervised prediction problem and the raw reconstruction decoder: frozen latent-field
prediction for CS-JEPA versus raw future-field reconstruction for \recon{}.  Thus, the target
family is the substantive difference, while the reference receives 9,607 additional training-only
parameters.  The scope is correspondingly precise: this tests the collective prediction objective
above a shared frozen JEPA-pretrained local encoder, not two representation pipelines trained
independently from scratch.  A registered crossed experiment separately replaces that shared
encoder with matched reconstruction-pretrained checkpoints, testing whether the stage-1 advantage
is merely inherited from a JEPA-aligned stage~0 objective.

Several tempting shortcuts would make the task look decentralized while leaking the answer.
Downstream collective labels are therefore hidden during representation learning.  Label-subset
selection is episode-level, stratified by task and training swarm size, nested within each draw,
and performed without reading label values.  Model-selection episodes choose one representation
checkpoint per method and seed but never fit the downstream probe.  ID, topology-OOD, and size-OOD evaluation
episodes are excluded from both checkpoint and probe fitting.  Finally, deployment audits call the
online prediction path without target-global, target-adjacency, target-active, or collective-label
tensors.  Together with zeroed clock and action channels, these checks rule out future-input
leakage, episode-progress identification, and centralized aggregation at test time.

\subsection{Label efficiency and endpoints}

For each outer seed, we train a ridge probe ($\alpha=10^{-3}$) on 6, 12, or 24 globally labeled
episodes.  Five fixed, task-and-size-stratified draws are nested within each budget; subset selection
does not inspect labels.  Accuracy is fixed-physical-scale normalized MSE over the ten decoded
quantities.  Agreement is the unnormalized MSE between decoded robot predictions within an episode.
Lower is better for both.  Because the two metrics have different scales, we interpret agreement
through paired within-metric effects and jointly with accuracy, not by comparing their numerical
magnitudes.

More explicitly, for a time window with $N$ active robots, decoded predictions
$\widehat y^i\in\mathbb R^{10}$, common target $y$, and fixed physical scales $s_d$, the
window contributions are
\begin{align}
E_{\mathrm{acc}}&=\frac{1}{10N}\sum_{i=1}^{N}\sum_{d=1}^{10}
\left(\frac{\widehat y_d^i-y_d}{s_d}\right)^2, \\
E_{\mathrm{agr}}&=\frac{1}{10N}\sum_{i=1}^{N}\sum_{d=1}^{10}
\left(\widehat y_d^i-\overline{\widehat y}_d\right)^2,
\quad \overline{\widehat y}=\frac{1}{N}\sum_i\widehat y^i.
\label{eq:endpoints}
\end{align}
The centroid in $E_{\mathrm{agr}}$ is used only to score dispersion after inference; it is not fed
to any robot and is not the primary prediction.  Window values are reduced to episode values and
then task-balanced, so large swarms and long episodes do not silently receive more inferential
weight.

No single label budget should determine whether a representation is called label-efficient.  The
primary endpoint therefore summarizes the full learning curve with a log-budget-span-normalized
trapezoidal AUC,
$\mathrm{AUC}=0.25E_6+0.50E_{12}+0.25E_{24}$, where $E_k$ is the error at
label budget $k$.
We report the paired effect
$\Delta=\mathrm{AUC}_{\recon}-\mathrm{AUC}_{\candidate}$, so positive values favor CS-JEPA.
Episode is the within-seed resampling unit and training seed the outer unit ($n=5$ initially;
$n=12$ per extension).  We report 10,000-resample outer-seed bootstrap intervals and favorable-seed
counts.  Episodes are task-stratified; methods share label draws and evaluation data.  Only
independently trained seeds enter outer uncertainty, and the two cohorts are analyzed separately.
Split families, budgets, the $\lambda=2$ anchor, and full-curve AUC were fixed before
evaluation, avoiding post-hoc budget selection.

\subsection{Fresh-cohort replication and objective-isolation controls}

To test whether the initial pattern survives new training randomness and data, the fresh
replication uses a new cohort and twelve new stage-1 seeds while retaining the stage-0 checkpoint,
splits, budgets, draws, architecture, and endpoints.  We analyze it separately from the initial
cohort.  Shared absolute position is an obvious alternative explanation, so a second experiment
retrains both stages with self-relative receiver positions, a centroid-centered teacher field, and
zeroed task-vector channels.  Shared orientation, metric scale, graph, and message budget remain;
the ablation therefore tests translation rather than full frame invariance.  Both experiments use
paired data and exact outer-seed tests.

We also cross the stage~0 and stage~1 objectives in a registered $2\times2$ control.  Stage~0 uses
either latent-mean JEPA prediction or matched raw-future reconstruction, while stage~1 uses either
CS-JEPA or \recon{}.  Eight independently trained stage~0 seeds are paired across all four cells;
the stage~1 initialization is fixed, and all data, architectures, endpoints, and selection rules
remain unchanged.  The primary endpoint is the equal-weighted four-split accuracy AUC contrast
between stage~1 objectives under reconstruction-pretrained stage~0.  The JEPA-pretrained contrast
and encoder-by-objective interaction are supporting endpoints.  Stage~0 training seed is the outer
unit, with an exact paired sign-flip test.  Because the data are fixed, this is an objective-isolation
experiment rather than another fresh-data replication.

\subsection{Action-conditioned counterfactual value estimation}

Prediction becomes useful for planning only if it distinguishes the consequences of available
actions.  We therefore train matched action-conditioned variants with eight outer seeds.  Each
four-step plan conditions \collectiveprediction{} before a frozen predictor and scalar ridge
readout estimate its outcome.  Both methods share the stage-0 encoder, receiver-local architecture,
communication, anchor, plans, and counterfactual data.  Each receiver maps its conditioned
representation and own plan to
$u(a)=\text{task}-0.5(1-\text{connectivity})-2\,\text{collision}$.  Value MSE is
computed on $u$.  Pearson correlation and regret use
$s(a)=u(a)-0.01\,\frac{1}{4}\sum_{\tau=1}^{4}\alpha_\tau^2-0.02|\alpha_1|$;
the predicted score replaces $u(a)$ by $\widehat u(a)$ while retaining the same known costs.
The readout uses 1,800 branches from 12 base episodes and a disjoint test set of 1,800 branches,
12 new base episodes, 72 shared contexts, and $N\in\{8,16,32\}$; robots are not averaged.

This exhaustive branch supervision asks whether the representation contains planning-relevant
information; it is separate from the label-efficiency claim.  Paired value MSE is
primary; Pearson correlation is computed within each context--receiver group over 25 plans and
averaged within seed.  Training seed is the outer unit ($n=8$), with an exact paired sign-flip test
and a 100,000-resample bootstrap interval.

\subsection{Fully decentralized dual-axis rigid-body control}

Better offline value estimates do not automatically produce better behavior, so the final study
closes the loop.  Across 16 outer seeds, fresh matched action-conditioned predictors share the
frozen encoder, recurrent backbone, anchor, communication, and deployment capacity.  Same-capacity receiver-local
heads use only counterfactual branches---never closed-loop outcomes---with identical listwise,
0.25-weighted score-regression, and 2.0-weighted class-balanced plan losses.  Cross-receiver
consistency has zero weight.

Each robot independently scores a $5\times5$ cohesion--cruise grid over a 0.8-s horizon, executes
four steps, and replans.  Predictions, scores, and plans are never pooled; communication remains
the 256-byte recurrent message, with no centralized selector.

The nominal comparator executes the fixed flocking law directly, with alignment, cohesion,
separation, and cruise gains $(0.9,0.12,1.8,0.55)$, and uses no learned prediction or plan
selection.  The matched \recon{} comparator uses the same candidate grid, value-head capacity, and
receiver-local planning procedure as CS-JEPA.

Evaluation uses Crazyflie \texttt{cf2x} rigid-body PyBullet dynamics with drag, 240-Hz physics,
40-Hz control, and a 0.1-s model step.  We test $N\in\{8,16\}$ under clean communication, 20\%
packet loss, one-step delay with jitter, and combined loss--delay--jitter with asynchronous updates.
Four episodes per size and condition yield 32 paired episodes per outer seed; episode seeds and
faults are shared across methods.  Composite utility is
$u=\text{task}-0.5(1-\text{connectivity})-2\,\text{collision}$.  Training seed is the outer unit;
inference uses exact paired sign-flip tests and 100,000-resample outer-seed intervals.

%%%%%%%%%%%%%%%%%%%%%%%%%%%%%%%%%%%%%%%%%%%%%%%%%%%%%%%%%%%%%%%%%%%%%%%%%%%%%%%%
\section{Results}

\subsection{Compatible predictions emerge without consensus training}

The central fresh-cohort result is joint.  With zero weight on any agreement term, CS-JEPA reduces
decoded receiver-to-receiver disagreement relative to \recon{} on every evaluated split and in all
12 independently trained stage-1 seeds (Table~\ref{tab:replication}).  Robots neither exchange
predictions nor receive a consensus target constructed from their outputs.  Compatibility therefore
emerges from the shared target semantics rather than from output pooling or an explicit consensus
penalty.

\begin{figure*}[t]
\centering
\begin{tikzpicture}[font=\footnotesize]
% Agreement panel: raw values are mapped with 75 cm per unit.
\begin{scope}[xshift=0cm]
\node[font=\bfseries] at (3.00,2.28) {Agreement AUC improvement};
\node[text=neutralgray] at (3.00,1.91) {$\Delta=0.04631\;[0.03893,0.05507]$};
\draw[->,draw=neutralgray!85,line width=.55pt] (0,0) -- (6.35,0);
\foreach \x/\lab in {0/0,1.5/.02,3.0/.04,4.5/.06,6.0/.08}{
  \draw[draw=neutralgray!65] (\x,-.06) -- (\x,.06);
  \node[anchor=north,text=neutralgray] at (\x,-.10) {\lab};
}
\draw[draw=neutralgray!35,dashed] (0,.18) -- (0,1.70);
\draw[draw=targetorange,line width=1.2pt] (2.920,1.50) -- (4.130,1.50);
\draw[draw=targetorange,line width=.8pt] (2.920,1.39) -- (2.920,1.61);
\draw[draw=targetorange,line width=.8pt] (4.130,1.39) -- (4.130,1.61);
\node[diamond,draw=targetorange,fill=white,minimum size=6.4pt,inner sep=0pt,line width=1pt]
  at (3.473,1.50) {};
\foreach \x/\y in {2.805/.34,3.439/.58,3.285/.82,5.853/1.06,2.724/.34,2.767/.58,
                    2.434/.82,3.667/1.06,3.772/.34,2.111/.58,3.414/.82,5.412/1.06}{
  \fill[targetorange!92] (\x,\y) circle (1.85pt);
}
\node[anchor=north,text=neutralgray] at (3.00,-.48) {CS-JEPA advantage in disagreement AUC};
\end{scope}
% Accuracy panel: raw values are mapped with 140 cm per unit.
\begin{scope}[xshift=8.55cm]
\node[font=\bfseries] at (3.15,2.28) {Accuracy AUC improvement};
\node[text=neutralgray] at (3.15,1.91) {$\Delta=0.03046\;[0.02623,0.03441]$};
\draw[->,draw=neutralgray!85,line width=.55pt] (0,0) -- (6.55,0);
\foreach \x/\lab in {0/0,1.4/.01,2.8/.02,4.2/.03,5.6/.04}{
  \draw[draw=neutralgray!65] (\x,-.06) -- (\x,.06);
  \node[anchor=north,text=neutralgray] at (\x,-.10) {\lab};
}
\draw[draw=neutralgray!35,dashed] (0,.18) -- (0,1.70);
\draw[draw=jepablue,line width=1.2pt] (3.672,1.50) -- (4.818,1.50);
\draw[draw=jepablue,line width=.8pt] (3.672,1.39) -- (3.672,1.61);
\draw[draw=jepablue,line width=.8pt] (4.818,1.39) -- (4.818,1.61);
\node[diamond,draw=jepablue,fill=white,minimum size=6.4pt,inner sep=0pt,line width=1pt]
  at (4.265,1.50) {};
\foreach \x/\y in {2.802/.34,5.129/.58,4.724/.82,5.343/1.06,3.878/.34,4.407/.58,
                    3.607/.82,3.095/1.06,4.945/.34,2.548/.58,5.601/.82,5.099/1.06}{
  \fill[jepablue!88] (\x,\y) circle (1.85pt);
}
\node[anchor=north,text=neutralgray] at (3.15,-.48) {CS-JEPA advantage in error AUC};
\end{scope}
\end{tikzpicture}
\caption{\textbf{Fresh twelve-seed replication at the outer-unit level.}  Each dot is one
independently trained stage-1 seed after equal weighting of the four evaluation families; diamonds
and whiskers are means and 95\% outer-seed bootstrap intervals.  Left: compatibility without
agreement training.  Right: corresponding accuracy evidence against collapse.  Positive favors
CS-JEPA; both effects are positive in 12/12 seeds (exact $p=0.000488$).}
\label{fig:fresh-effects}
\end{figure*}
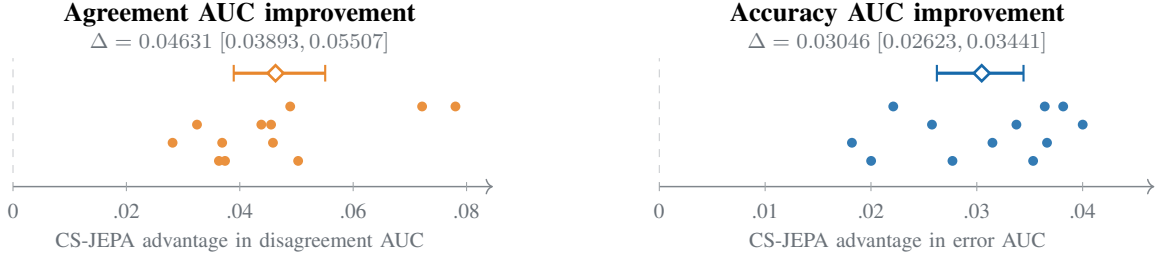

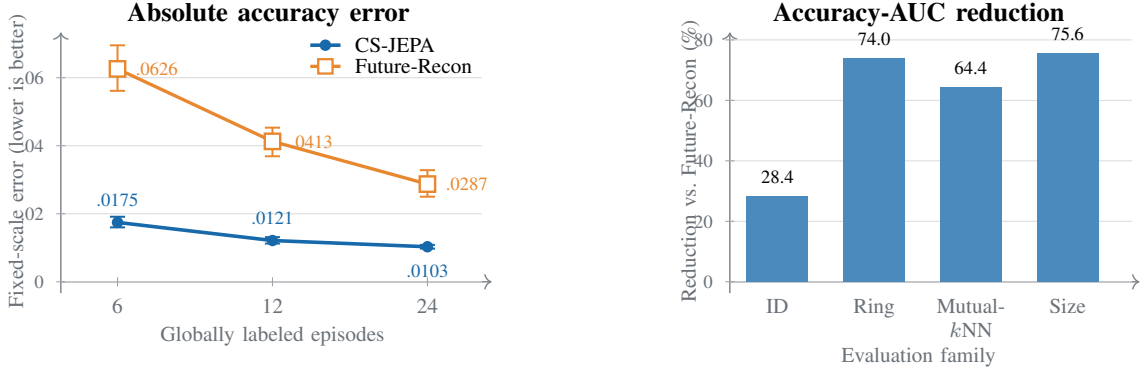
\begin{figure*}[t]
\centering
\begin{tikzpicture}[font=\footnotesize]
% Absolute equal-weighted curve.  Vertical scale is 45 cm per unit with a 0.35 cm baseline.
\begin{scope}[xshift=.25cm]
\node[font=\bfseries] at (3.30,3.88) {Absolute accuracy error};
\draw[->,draw=neutralgray!85,line width=.55pt] (.55,.35) -- (6.25,.35);
\draw[->,draw=neutralgray!85,line width=.55pt] (.55,.20) -- (.55,3.65);
\foreach \y/\lab in {.35/0,1.25/.02,2.15/.04,3.05/.06}{
  \draw[draw=neutralgray!25] (.55,\y) -- (6.12,\y);
  \draw[draw=neutralgray!65] (.49,\y) -- (.61,\y);
  \node[anchor=east,text=neutralgray,font=\scriptsize] at (.45,\y) {\lab};
}
\foreach \x/\lab in {1.30/6,3.35/12,5.40/24}{
  \draw[draw=neutralgray!65] (\x,.29) -- (\x,.41);
  \node[anchor=north,text=neutralgray] at (\x,.25) {\lab};
}
\node[anchor=north,text=neutralgray] at (3.35,-.12) {Globally labeled episodes};
\node[rotate=90,text=neutralgray] at (-.02,1.95) {Fixed-scale error (lower is better)};

% Future-Recon means and 95% outer-seed intervals.
\draw[draw=targetorange,line width=1.25pt] (1.30,3.1665) -- (3.35,2.2075) -- (5.40,1.6428);
\foreach \x/\lo/\hi in {1.30/2.8764/3.4784,3.35/2.0104/2.3889,5.40/1.4757/1.8267}{
  \draw[draw=targetorange,line width=.75pt] (\x,\lo) -- (\x,\hi);
  \draw[draw=targetorange,line width=.75pt] (\x-.10,\lo) -- (\x+.10,\lo);
  \draw[draw=targetorange,line width=.75pt] (\x-.10,\hi) -- (\x+.10,\hi);
}
\node[draw=targetorange,fill=white,minimum size=5.8pt,inner sep=0pt,line width=.9pt]
  at (1.30,3.1665) {};
\node[draw=targetorange,fill=white,minimum size=5.8pt,inner sep=0pt,line width=.9pt]
  at (3.35,2.2075) {};
\node[draw=targetorange,fill=white,minimum size=5.8pt,inner sep=0pt,line width=.9pt]
  at (5.40,1.6428) {};
\node[anchor=west,text=targetorange,font=\scriptsize] at (1.42,3.1665) {.0626};
\node[anchor=west,text=targetorange,font=\scriptsize] at (3.47,2.2075) {.0413};
\node[anchor=west,text=targetorange,font=\scriptsize] at (5.52,1.6428) {.0287};

% CS-JEPA means and 95% outer-seed intervals.
\draw[draw=jepablue,line width=1.35pt] (1.30,1.1358) -- (3.35,.8959) -- (5.40,.8135);
\foreach \x/\lo/\hi in {1.30/1.0700/1.2106,3.35/.8559/.9406,5.40/.7889/.8384}{
  \draw[draw=jepablue,line width=.75pt] (\x,\lo) -- (\x,\hi);
  \draw[draw=jepablue,line width=.75pt] (\x-.10,\lo) -- (\x+.10,\lo);
  \draw[draw=jepablue,line width=.75pt] (\x-.10,\hi) -- (\x+.10,\hi);
}
\fill[jepablue] (1.30,1.1358) circle (2.15pt);
\fill[jepablue] (3.35,.8959) circle (2.15pt);
\fill[jepablue] (5.40,.8135) circle (2.15pt);
\node[anchor=south,text=jepablue,font=\scriptsize] at (1.30,1.225) {.0175};
\node[anchor=south,text=jepablue,font=\scriptsize] at (3.35,.982) {.0121};
\node[anchor=north,text=jepablue,font=\scriptsize] at (5.40,.700) {.0103};

\draw[draw=jepablue,line width=1.35pt] (3.85,3.48) -- (4.25,3.48);
\fill[jepablue] (4.05,3.48) circle (2.15pt);
\node[anchor=west] at (4.32,3.48) {CS-JEPA};
\draw[draw=targetorange,line width=1.25pt] (3.85,3.20) -- (4.25,3.20);
\node[draw=targetorange,fill=white,minimum size=5.4pt,inner sep=0pt,line width=.85pt]
  at (4.05,3.20) {};
\node[anchor=west] at (4.32,3.20) {Future-Recon};
\end{scope}

% Family-wise relative AUC reductions.
\begin{scope}[xshift=9.15cm]
\node[font=\bfseries] at (3.00,3.88) {Accuracy-AUC reduction};
\draw[->,draw=neutralgray!85,line width=.55pt] (.50,.35) -- (5.92,.35);
\draw[->,draw=neutralgray!85,line width=.55pt] (.50,.20) -- (.50,3.65);
\foreach \y/\lab in {.35/0,1.15/20,1.95/40,2.75/60,3.55/80}{
  \draw[draw=neutralgray!25] (.50,\y) -- (5.82,\y);
  \draw[draw=neutralgray!65] (.44,\y) -- (.56,\y);
  \node[anchor=east,text=neutralgray,font=\scriptsize] at (.40,\y) {\lab};
}
\node[rotate=90,text=neutralgray] at (-.02,1.95) {Reduction vs. Future-Recon (\%)};
\fill[jepablue!78] (.72,.35) rectangle (1.54,1.4846);
\fill[jepablue!78] (2.00,.35) rectangle (2.82,3.3102);
\fill[jepablue!78] (3.28,.35) rectangle (4.10,2.9259);
\fill[jepablue!78] (4.56,.35) rectangle (5.38,3.3741);
\node[anchor=south,font=\scriptsize] at (1.13,1.535) {28.4};
\node[anchor=south,font=\scriptsize] at (2.41,3.360) {74.0};
\node[anchor=south,font=\scriptsize] at (3.69,2.976) {64.4};
\node[anchor=south,font=\scriptsize] at (4.97,3.424) {75.6};
\node[anchor=north,text=neutralgray] at (1.13,.25) {ID};
\node[anchor=north,text=neutralgray] at (2.41,.25) {Ring};
\node[anchor=north,text=neutralgray,align=center] at (3.69,.25) {Mutual-\\$k$NN};
\node[anchor=north,text=neutralgray] at (4.97,.25) {Size};
\node[anchor=north,text=neutralgray] at (3.00,-.40) {Evaluation family};
\end{scope}
\end{tikzpicture}
\caption{\textbf{Fresh-cohort label efficiency.}  Left: equal-weighted four-family absolute
accuracy error at 6/12/24 globally labeled episodes.  Points are means over 12 outer training
seeds; whiskers are 95\% outer-seed bootstrap intervals.  Right: relative log-budget AUC reduction
by family; every paired family-wise AUC effect favors CS-JEPA in 12/12 seeds.  The curves are
descriptive pointwise summaries; the preregistered primary inference remains the paired AUC
contrast in Table~\ref{tab:replication}.}
\label{fig:label-efficiency}
\end{figure*}

Agreement alone could be produced by collapse.  Here scale-normalized decoding accuracy improves
on the same seeds and splits: relative to \recon{}, accuracy AUC falls by 28.4\% on ID and by
64.4--75.6\% under topology and size shift.  The independently formed predictions are therefore
both more compatible and more informative.  Because the receiver anchor, frozen encoder, and
deployment path are matched, the precise supported contrast is learned latent-target prediction
against matched raw-future reconstruction.  Figure~\ref{fig:fresh-effects} shows the paired
outer-seed effects behind this fresh-cohort aggregate.
Figure~\ref{fig:label-efficiency} exposes the corresponding absolute learning curve: at
6/12/24 globally labeled episodes, equal-weighted four-family accuracy error is
.0175/.0121/.0103 for CS-JEPA and .0626/.0413/.0287 for \recon{}.

\begin{table}[t]
\caption{Fresh-cohort joint replication.  Values are reconstruction-minus-JEPA error AUC with
95\% outer-seed CIs; positive favors CS-JEPA.  Every row and metric is positive in 12/12 seeds
($p=0.000488$).}
\label{tab:replication}
\centering
\resizebox{\columnwidth}{!}{%
\begin{tabular}{lcc}
\toprule
Split & Accuracy $\Delta$ [95\% CI] & Agreement $\Delta$ [95\% CI] \\
\midrule
ID & .001568 [.001414, .001709] & .000981 [.000862, .001093] \\
Ring OOD & .048385 [.040558, .056713] & .076685 [.061555, .095064] \\
Mutual-$k$NN & .034180 [.028687, .039756] & .045911 [.038218, .053854] \\
Size OOD & .037714 [.031207, .044368] & .061672 [.049559, .076044] \\
\bottomrule
\end{tabular}
}
\end{table}

\noindent\textit{Direct-supervision trade-off.}
A separate deployment-matched five-seed diagnostic compares frozen CS-JEPA with capacity-matched
recurrent local GNN and attention predictors.  CS-JEPA is more accurate on all four families:
supervised-minus-CS-JEPA accuracy-error AUC ranges from $+0.074240$ to $+0.092186$, with every
95\% outer-seed CI above zero.  Direct supervision yields lower disagreement: the corresponding
contrast ranges from $-0.020292$ to $-0.000279$, with every interval below zero.  Thus direct
supervision can align predictions more tightly, whereas CS-JEPA provides substantially more
accurate physical forecasts while improving agreement over \recon{} without an agreement loss.

Without a shared absolute origin or task displacement, equal-weighted accuracy AUC falls from
0.008852 to 0.007248 (18.1\%; paired $\Delta=0.001604$ [0.001034, 0.002487]) and agreement AUC
from 0.005735 to 0.002729 (52.4\%; $\Delta=0.003006$ [0.001926, 0.004854]), both in 12/12 seeds
($p=0.000488$); every split and task aggregate favors CS-JEPA.

\noindent\textit{Crossed stage-0 objective control.}
The advantage survives when stage~0 itself is trained by raw-future reconstruction
(Table~\ref{tab:crossed-stage0}).  All four split estimates and all eight paired stage~0 seeds are
positive.  The supporting JEPA-pretrained contrast is also positive, while the registered
JEPA-minus-reconstruction interaction is $-0.00789$ [$-0.01241$, $-0.00326$]
($p=0.03125$), meaning that the CS-JEPA advantage is larger, not smaller, after reconstruction
pretraining.  Agreement improves in 8/8 seeds under either stage~0 objective.  The result therefore
isolates the stage~1 latent-target objective from a purely JEPA-aligned stage~0 explanation.

\begin{table}[t]
\caption{Crossed stage-0 objective control ($n=8$ paired stage-0 seeds; fixed stage-1
initialization).  $\Delta$ is equal-weighted accuracy error-AUC improvement; positive favors
CS-JEPA.}
\label{tab:crossed-stage0}
\centering
\scriptsize
\setlength{\tabcolsep}{2.5pt}
\begin{tabular}{lccc}
\toprule
Stage-0 objective & $\Delta$ [95\% CI] & Fav. & $p$ \\
\midrule
Latent-mean JEPA & +.02251 [.01987, .02511] & 8/8 & .0078125 \\
Raw-future reconstruction & +.03040 [.02455, .03693] & 8/8 & .0078125 \\
\bottomrule
\end{tabular}
\end{table}

\subsection{Action-conditioned value estimation}

\begin{table}[t]
\caption{Eight-seed four-step value comparison.  Lower MSE is better; positive
within-context Pearson $\Delta$ favors CS-JEPA.  Every MSE and correlation effect favors CS-JEPA in 8/8 seeds
($p=0.0078125$ for every row and metric).}
\label{tab:value-probe}
\centering
\scriptsize
\setlength{\tabcolsep}{3.0pt}
\resizebox{\columnwidth}{!}{%
\begin{tabular}{lcccc}
\toprule
Scope & CS-JEPA MSE & \recon{} MSE & Reduction & \shortstack{Within-context\\Pearson $\Delta$} \\
\midrule
Overall & .01553 & .02847 & 45.46\% & +.12907 \\
$N=8$ & .02011 & .04286 & 53.07\% & +.13967 \\
$N=16$ & .01418 & .01768 & 19.77\% & +.12105 \\
Unseen $N=32$ & .01228 & .02487 & 50.62\% & +.13044 \\
\bottomrule
\end{tabular}
}
\end{table}

CS-JEPA improves value MSE and within-context score correlation at both training sizes and unseen
$N=32$ (Table~\ref{tab:value-probe}), confirming action-relevant information beyond the decoded
collective variables.

\subsection{Latent-target prediction improves fully decentralized closed-loop utility}

Across clean and communication-fault episodes, CS-JEPA improves utility over matched
reconstruction by 5.3\%, favorable in 12/16 seeds.  Task score and connectivity move in the same
direction, and no method has a collision or near-collision in the evaluated episodes.  Against
nominal control, utility, task score, and connectivity improve in every seed
(Table~\ref{tab:closed-loop}).  Here each robot chooses from its own prediction, so the result links
the training target to realized decentralized behavior rather than only a post-hoc score.

\begin{table}[t]
\caption{Fully decentralized dual-axis rigid-body control ($n=16$).  $\Delta$ is CS-JEPA minus
comparator; higher is better.  CIs and exact tests use training seed as the outer unit.}
\label{tab:closed-loop}
\centering
\scriptsize
\setlength{\tabcolsep}{2.2pt}
\resizebox{\columnwidth}{!}{%
\begin{tabular}{llccc}
\toprule
Outcome & Comparator & $\Delta$ [95\% CI] & Fav. & $p$ \\
\midrule
Composite utility & \recon{} & +.02088 [.00490, .03606] & 12/16 & .02295 \\
Task score & \recon{} & +.01344 [.00311, .02324] & 11/16 & .02371 \\
Connectivity & \recon{} & +.01487 [.00348, .02550] & 11/16 & .02197 \\
Composite utility & Nominal & +.03324 [.02172, .04481] & 16/16 & .000031 \\
Task score & Nominal & +.02124 [.01381, .02882] & 16/16 & .000031 \\
Connectivity & Nominal & +.02401 [.01596, .03204] & 16/16 & .000031 \\
\bottomrule
\end{tabular}
}
\end{table}

A complementary single-axis receiver-local study shows that the latent target also aligns
independent decisions without a consistency loss: all-receiver first-action agreement increases by
0.14472 [0.10282, 0.18618] and pairwise disagreement decreases by 0.05888 [0.04048, 0.07701]
relative to \recon{}, with both effects favorable in 15/16 seeds.

%%%%%%%%%%%%%%%%%%%%%%%%%%%%%%%%%%%%%%%%%%%%%%%%%%%%%%%%%%%%%%%%%%%%%%%%%%%%%%%%
\newpage
\section{Discussion and Limitations}

Compatible receiver-local predictions emerge without consensus training.  Each robot sees
different evidence, keeps a different recurrent state, and never reads another robot's prediction,
yet decoded disagreement falls on fresh seeds and every evaluated split.  This would be
uninteresting if the robots converged on a constant; instead, physical decoding accuracy improves
on the same comparisons.  That joint result makes agreement at zero agreement-loss weight evidence
for a useful shared representation.

The mechanism is correspondingly precise.  CS-JEPA does not simply ask robots to predict a common
future---the matched reference also reconstructs one shared raw future.  It asks them to predict a
learned latent target that can retain collective structure while discarding raw details that need
not be identical across local views.  With message width, deployment capacity, receiver anchor,
and the absence of output pooling held fixed, the comparison isolates that target choice.  Matched
64-D $h_t$ probes place the compatibility effect before the final readout.

The transfer results clarify what information makes this possible.  Removing absolute origin and
task displacement weakens an obvious coordinate shortcut, yet the advantage remains; shared
orientation and scale are still assumed.  The crossed stage~0 control goes further: replacing the
JEPA-pretrained encoder with a reconstruction-pretrained counterpart preserves and enlarges the
stage~1 advantage.  The gain therefore comes from latent-target prediction rather than simply
inheriting a JEPA-aligned stage~0 encoder.  Scaling to larger swarms does not require wider messages or
more tokens.  Instead, the recurrent history provides a finite multi-hop path through which
evidence can travel, and every evaluated large-swarm window is reachable within that history.

The value and control studies then ask whether this compatibility matters after prediction.
Improved counterfactual value estimates show that the latent future retains action-relevant
structure, and the rigid-body study shows that each robot can use that structure through the same
receiver-local interface under clean and faulty communication.

\noindent\textit{Scope of the evidence.}
The evidence is simulator-only, retains shared orientation and scale, and covers the reported graph
and size families; hardware transfer, independent local frames, and specialized controllers remain
open.  Improved value estimation does not reduce selected-plan regret.  Privileged future state
and exhaustive branch labels are used only during
training.  These boundaries restrict the claim to compatible receiver-local prediction and its
demonstrated label-efficient and downstream utility.

%%%%%%%%%%%%%%%%%%%%%%%%%%%%%%%%%%%%%%%%%%%%%%%%%%%%%%%%%%%%%%%%%%%%%%%%%%%%%%%%
\section{Conclusion}

CS-JEPA does not train robots to agree; it trains each robot to predict the same latent future.
Across fresh seeds, topology shifts, and swarm-size shifts, those independent predictions become
more compatible even though agreement is absent from the loss.  Their accuracy improves at the
same time, showing that compatibility does not come from collapse, and low-label probes show that
the shared state remains readily accessible.  Translation-free and crossed-objective controls
weaken coordinate and pretraining alternatives, while action-conditioned value estimation and
independently executed control connect the representation to decisions.  A shared latent future
thus aligns fragmented local evidence without pooling predictions or plans.

%%%%%%%%%%%%%%%%%%%%%%%%%%%%%%%%%%%%%%%%%%%%%%%%%%%%%%%%%%%%%%%%%%%%%%%%%%%%%%%%
\par\noindent\begin{minipage}[t]{\columnwidth}
\section*{Acknowledgment}

OpenAI ChatGPT assisted language editing under author supervision.
\end{minipage}

\balance
\bibliographystyle{vendor/IEEEtran}
\bibliography{references}

\end{document}